\documentclass[runningheads]{llncs}
\usepackage[colorlinks,linkcolor=blue,anchorcolor=blue,citecolor=blue]{hyperref}
\usepackage{graphicx}
\usepackage{amsmath}
\usepackage{array}
\usepackage{booktabs}
\usepackage{cite}
\usepackage[misc,geometry]{ifsym}
\newcolumntype{N}{>{\centering\arraybackslash}m{.5in}}
\newcolumntype{G}{>{\centering\arraybackslash}m{2in}}
\begin{document}
\title{Context-Aware Refinement Network Incorporating Structural Connectivity Prior for Brain Midline Delineation}
\titlerunning{Structural Connectivity Prior for Brain Midline Delineation}
%
\author{
Shen Wang\inst{1,2}\thanks{This work was done when Shen Wang was an intern at Deepwise AI Lab.} \and
Kongming Liang\inst{3}$^{\left(\textrm{\Letter}\right)}$ \and
Yiming Li\inst{4} \and \\
Yizhou Yu\inst{4} \and
Yizhou Wang\inst{2,3,5}
}

%
\authorrunning{S. Wang et al.}

%
\institute{
{Center for Data Science, Peking University, Beijing, China} \and
{Advanced Institute of Information Technology, Peking University, Hangzhou, China} \and
{Department of  Computer Science, Peking University, Beijing, China} \and
{Deepwise AI Lab, Beijing, China} \and
{Center on Frontiers of Computing Studies, Peking University, Beijing, China}
}
\maketitle              

\begin{abstract}
Brain midline delineation can facilitate the clinical evaluation of brain midline shift, which plays an important role in the diagnosis and prognosis of various brain pathology. Nevertheless, there are still great challenges with brain midline delineation, such as the largely deformed midline caused by the mass effect and the possible morphological failure that the predicted midline is not a connected curve. To address these challenges, we propose a context-aware refinement network (CAR-Net) to refine and integrate the feature pyramid representation generated by the UNet. Consequently, the proposed CAR-Net explores more discriminative contextual features and larger receptive field, which is of great importance to predict largely deformed midline. For keeping the structural connectivity of the brain midline, we introduce a novel connectivity regular loss (CRL) to punish the disconnectivity between adjacent coordinates. Moreover, we address the ignored prerequisite of previous regression-based methods that the brain CT image must be in the standard pose. A simple pose rectification network is presented to align the source input image to the standard pose image. Extensive experimental results on the CQ dataset and one inhouse dataset show that the proposed method requires fewer parameters and outperforms three state-of-the-art methods in terms of four evaluation metrics. Code is available at \url{https://github.com/ShawnBIT/Brain-Midline-Detection}.


\keywords{Brain midline delineation \and Computer aided diagnosis \and Context-aware refinement network \and Connectivity regular loss}
\end{abstract}

\section{Introduction}
The human brain in healthy subjects is approximately bilateral symmetrical and divided into two cerebral hemispheres that are separated by the ideal midline on the axial plane of CT images. However, various pathological conditions, such as traumatic brain injuries, strokes, and tumors, could break the symmetry by distorting the ideal midline (IML) to deformed midline (DML) and lead to brain midline shift (MLS). As a sign of increased intracranial pressure, the degree of MLS can serve as a quantitative indicator for physicians to make diagnosis and outcome prediction more accurate. For example, the guideline of Brain Trauma Foundation recommended emergency surgery for any traumatic epidural, subdural, or intracerebral hematoma causing an MLS larger than 5 mm~\cite{liao2018brain}. Since the complex and quantitative analysis of MLS is challenging and time-consuming for neurologists, computer-aided brain midline delineation could not only improve the accuracy and efficiency of MLS estimation~\cite{wei2019regression} but also reduce the interrater variability among neurologists~\cite{pisov2019incorporating}.

Traditional methods for brain midline delineation are classified into two types: symmetry-based~\cite{liao2006tracing, chen2015automatic} and landmark-based ones~\cite{liu2014automatic, chen2010actual}. For example, Liao et al.~\cite{liao2006tracing} decomposed the deformed midline into three segments and formulated the central curved segment as a quadratic Bezier curve, which is fit by using local symmetry. Liu et al.~\cite{liu2014automatic} proposed to build the deformed midline by localizing the anatomical points. However, these traditional methods may fail in the cases with largely deformed brain due to the following two reasons: (1) The midline is relatively difficult to be identified given low soft-tissue contrast; (2) The predefined anatomical points or parts may not be visible due to large deformation.~\cite{wei2019regression}

Recently, approaches based on deeplearning~\cite{wei2019regression, pisov2019incorporating, wang2020segmentation} have served in brain midline delineation, which can overcome the above issues to some extent. Hao et al.~\cite{wei2019regression} formulated the brain midline delineation as a regression task and proposed a regression-based line detection network. Pisov et al.~\cite{pisov2019incorporating} introduced a two-head convolutional neural network with shared input layers to predict the midline limits and regress the midline coordinates. However, the performance of such regression-based methods is limited due to the following aspects: (1) They ignore the structural connectivity prior that the midline is a connected and smooth curve. (2) The feature extraction network is not well designed for a largely deformed midline, or harder to train due to the high complexity. (3) They all share a common assumption that for each vertical coordinate $y$ there is at most one horizontal coordinate $x$ of midline pixel, which may fail in some extreme poses of the brain. For taking the structural connectivity prior into account, Wang et al.~\cite{wang2020segmentation} proposed a post-processing stage called pathfinding based on the segmentation probability map to build the midline. Their method can not be trained end-to-end which is sub-optimal.

To address such issues, this paper proposes a context-aware refinement network (CAR-Net) to enhance the feature extraction ability and introduce a novel connectivity regular loss (CRL) to incorporate prior knowledge of midline structural connectivity. Specifically, the main contributions are summarized as follows: (1) We propose a context-aware refinement network (CAR-Net) to refine and integrate the base feature pyramid for exploring more discriminative contextual features and larger receptive field. (2) We introduce a novel connectivity regular loss (CRL) to model the connectivity prior explicitly and guarantee the connectivity of the predicted midline. (3) We address the prerequisite ignored by the previous regression-based method and present a simple pose rectification module to satisfy the above prerequisite. The proposed method is evaluated on the CQ dataset and one inhouse dataset with the results showing that our method outperforms three state-of-the-art methods with fewer parameters.

\begin{figure}[t]
\centering
\includegraphics[width=\textwidth]{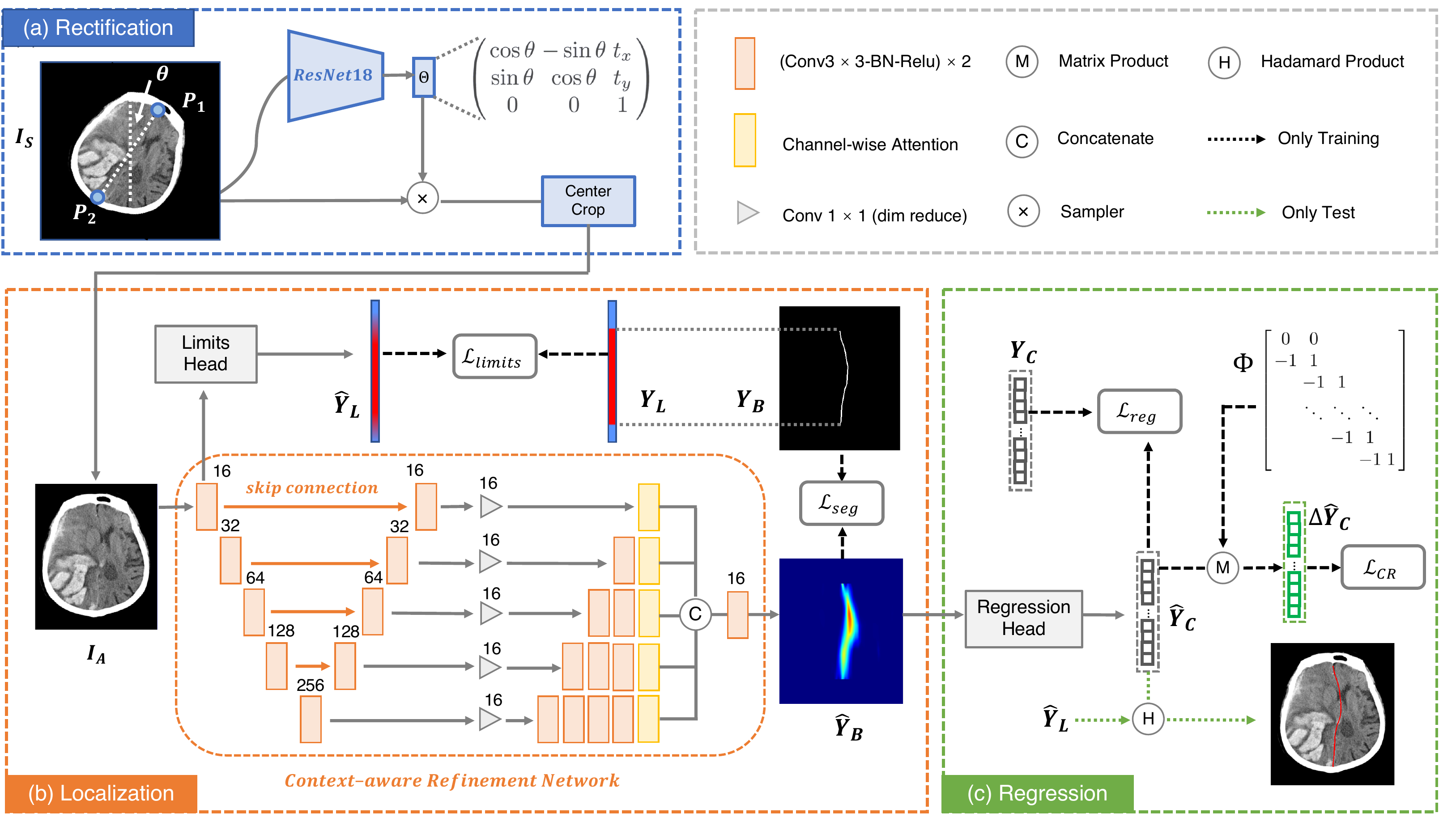}
\caption{The illustration of the pipeline of our proposed method for brain midline delineation, which consists of three parts, (a) rectification, (b) localization and (c) regression.}
\label{pipeline}
\end{figure}

\section{Method}
Fig.\ref{pipeline} shows the pipeline of our proposed method for brain midline delineation, which consists of three parts, (a)rectification, (b)localization and (c)regression. First, we present a pose rectification network to align the source CT image $I_{S}$ to a canonical pose image $I_{A}$. Second, the proposed context-aware refinement network (CAR-Net) takes the aligned CT image $I_{A} \in R^{H \times W}$ as input and generate the midline limits $\hat{Y}_{L} \in R^{H}$ (the vertical range of midline coordinates) through the limits head~\cite{pisov2019incorporating} and the segmentation probability map  $\hat{Y}_{B} \in R^{H \times W}$ of the midline band(the width expanded midline). Finally, the regression head~\cite{wei2019regression} takes the segmentation probability map $\hat{Y}_{B}$ as input and outputs the midline coordinates $\hat{Y}_\text{C} \in R^{H}$. In addition, the midline coordinates $\hat{Y}_\text{C}$ is multiplied by the transformation matrix $\Phi$ and obtain the adjacent coordinate difference vector $\Delta \hat{Y}_{C} \in R^{H}$, which can be utilized to compute the connectivity regular loss $L_{CR}$. We adopt the same structure of the limits head~\cite{pisov2019incorporating} and the regression head~\cite{wei2019regression}.


\subsection{Pose Rectification Module}
Previous methods share a common assumption that for each vertical axis coordinate $y$, there is at most one horizontal coordinate $x$ of midline pixel, which may fail in some extreme poses of the brain, due to improper distance, angle or displacement between the camera and patients, especially in real clinical application. Thus, we present a pose rectification network to align the images to the standard pose, which can guarantee the above assumption.

As shown in Fig.\ref{pipeline}(a), the anterior flax point $P_{1}$ and posterior flax point $P_{2}$ of the ground truth midline are used to calculate the rotational angle and the brain center, which can form as a rigid transformation. Then, we can align the source CT image $I_{S}$ to form the target image pair $I_{T}$. Given $I_{S}$, the pose rectification network $\phi$ transforms $I_{S}$ to $\phi(I_{S})$. Specifically,  we use a light-weighted ResNet-18~\cite{he2016deep} as the backbone of $\phi$ and minimize the loss $L_{2}(\phi(I_{S}), I_{T})$. The output of the pose rectification network is a group of parameters ($t_{x}$, $t_{y}$, $\theta$) of rigid transformation. $t_{x}$ and $t_{y}$ stand for horizontal and vertical displacements and $\theta$ stands for the rotational angle. To this end, $I_{S}$ is transformed to $I_{A}$ following:

\begin{equation}
\label{rectification}
I_{A} = \phi(I_{S})=B\left(\left(\begin{array}{ccc}
{ \cos\theta} & {\quad- \sin\theta} & {\quad t_{x}} \\
{ \sin\theta} & { \quad\cos\theta} & {\quad t_{y}}
\end{array}\right) G(I_{S}),\quad I_{S}\right)
\end{equation}
where $B$ stands for a bilinear interpolating function, and $G$ represents a regular grid function. Furthermore, the aligned images are center cropped to a uniform size for the midline delineation.

\subsection{Context-Aware Refinement Network}
For the midline delineation task, the normal parts of the midline are easy to process. However, it is difficult to locate the shifted parts of the largely deformed midline accurately, which requires a larger receptive field and more discriminative contextual information. As shown in ~\cite{zhang2018exfuse}, low-level and high-level features are complementary by nature, where low-level features are rich in spatial details and high-level features are rich in semantic concepts. Therefore, based on the feature pyramid representation $\{ f_{i}\mid i=1,2,3,4,5\} $ generated by U-Net~\cite{ronneberger2015u}, we attach a context-aware feature refinement module, which can refine each scale features and integrate them adaptively to explore more discriminative contextual features and achieve larger receptive field for the harder shifted parts of the deformed midline.

Specifically, as shown in Fig.\ref{pipeline}(b), we first refine each scale feature map $f_{i}$ to obtain local refined feature representation ${f_{i}}^{l}$ by applying several basic convolution blocks. Given the trade-off between effectiveness and efficiency, more basic convolution blocks are stacked into deeper layers. Then we adopt the SE block~\cite{hu2018squeeze} as the channel-wise attention, which can recalibrate the local refined representation ${f_{i}}^{l}$ to extract more discriminative features ${f_{i}}^{a}$ for a specific scale.
Finally, the representative features ${f_{i}}^{a}$ of different levels are integrated via bilinear interpolation upsampling, concatenating and one basic convolution block to form the context-aware refinement representation $f^{R}$.
Compared to the feature representation $f_{1}$ of original UNet, the context-aware refinement representation $f^{R}$ have larger receptive field and more discriminative contextual information.

\subsection{Connectivity Regular Loss}
For the supervision of the midline coordinates, the previous regression-based methods only used mean square error loss (MSE)~\cite{wei2019regression, pisov2019incorporating}. They ignored the structural connectivity prior that the brain midline is a continuous curve, which may lead to the possible discontinuity of the midline. The segmentation-based method~\cite{wang2020segmentation} proposed post-processing, which relies heavily on the segmentation probability map of the midline and cannot be optimized in an end-to-end way. Based on the above observations, we propose a novel continuity regular loss (CRL) to incorporate structural connectivity prior, which can keep the morphology consistency between the predicted midline and the ground truth midline.

Specifically, we first give the definition of the midline connectivity. For the midline coordinates $X=(x_{1}, x_{2}, ..., x_{n})^\mathrm{T}$, if $|x_{i} - x_{i-1}| \leq \delta$ holds for every $i = 2,3,...,n$, we call the the midline coordinates $X$ satisfy $\delta$-$connectivity$. Then we denote $\Delta X=(0, \Delta x_{1}, \Delta x_{2}, ..., \Delta x_{n})^\mathrm{T}$, where $\Delta x_{i}=x_{i} - x_{i-1}$ for every $i = 2,3,...,n$. The derivation between $X$ and $\Delta X$ are as follows:
\begin{equation}
\Delta X=\left[\begin{array}{c}
{0} \\
{x_{1}-x_{0}} \\
{x_{2}-x_{1}} \\
{\vdots} \\
{x_{n-1}-x_{n-2}} \\
{x_{n}-x_{n-1}}
\end{array}\right]=\left[\begin{array}{ccccc}
{0} & {0} & {} & {} & {} \\
{-1} & {1} & {} & {} & {} \\
{} & {-1} & {1} & {} & {} \\
{} & {\ddots} & {\ddots} & {\ddots} & {} \\
{} & {} & {-1} & {1} & {} \\
{} & {} & {} & {-1} & {1}
\end{array}\right]\left[\begin{array}{c}
{x_{0}} \\
{x_{1}} \\
{x_{2}} \\
{\vdots} \\
{x_{n-2}} \\
{x_{n-1}} \\
{x_{n}}
\end{array}\right]=\Phi X
\end{equation}
where $\Phi$ is the transformation matrix.
Thus, we define the CRL as follows, which can effectively punish the disconnectivity between adjacent coordinates with the margin $\delta$ to guarantee the predicted midline coordinates $\hat{Y}_{C}$ satisfy $\delta$-$connectivity$.
\begin{equation}\label{connect}
\setlength\abovedisplayskip{7pt}
\setlength\belowdisplayskip{7pt}
L_{C R}(\hat{Y}_{C}) = f(\Delta \hat{Y}_{C}) = f(\Phi \cdot \hat{Y}_{C}),\ \text{where} \,
f(\boldsymbol x) = \sum_{i=1}^{n} \max \left(0,|x_{i}|-\delta)\right)
\end{equation}

\subsection{Loss Function and Optimization}
The whole framework is trained in an end-to-end way except the pose rectification network. The loss function $\mathcal{L}_{\text {limits}}$ of midline limits $\hat{Y}_{L}$ is the binary cross entropy loss and the loss function $\mathcal{L}_{\text {seg}}$ of segmentation probability map $\hat{Y}_{B}$ is the weight cross entropy loss. For the supervision of midline coordinates $\hat{Y}_{C}$ , we take $L_{1}$ loss as the regression loss $\mathcal{L}_{\text {reg}}$ and connectivity regular loss $L_{CR}$ as the regular term. The total loss function of the midline delineation is defined as:
\begin{equation} \label{total} 
\mathcal{L}_{total}=\lambda \mathcal{L}_{\text {limits}}+\gamma \mathcal{L}_{\text {seg}} + \xi \mathcal{L}_{reg} + \mu \mathcal{L}_{CR}
\end{equation}
where $\lambda,\gamma, \xi$ and $\mu$ denote the balanced weights of different parts.

In the inference phase, the source input CT image $I_{S}$ is first aligned to the standard pose image $I_{A}$ by the pose rectification network. Then the aligned image $I_{A}$ is sent to the CAR-Net and regression head successively to obtain the midline limits $\hat{Y}_{L}$ and the midline coordinates $\hat{Y}_{C}$. The midline limits $\hat{Y}_{L}$ are converted to binary one by a suitable threshold. Then the midline coordinates $\hat{Y}_{C}$ is multiplied by the midline limits $\hat{Y}_{L}$ with Hadamard product to form the real midline coordinates. Finally, we draw the real midline coordinates into the aligned image $I_{A}$, as shown in Fig.~\ref{pipeline}(c).

\section{Experiments}

\subsubsection{Dataset and Evaluation Metric.} We evaluate our method on the CQ dataset and one inhouse dataset. The CQ dataset is a subset of CQ500 dataset\footnote{\url{http://headctstudy.qure.ai/dataset}}, which consists of 63 midline shift subjects and the same number of healthy subjects. 59\% of the subjects have a significant midline shift ($\geq$ 5mm) and the mean MLS is $7.59 \pm 5.16$mm. Our inhouse dataset consists of 203 CT series which have different degrees of MLS caused by cerebral hemorrhage. 78\% of the subjects have a significant midline shift ($\geq$ 5mm) and the mean MLS is $9.04 \pm 5.54$mm. For both datasets, a total of 10 CT slices with the largest brain area in each subject were selected to be manually delineated by doctors for the midline golden standard. For the CQ dataset and our inhouse dataset, we randomly split the dataset into 76/20/30 and 120/30/53 as train/validation/test set respectively. We employ four metrics to measure the midline delineated by different methods, including line distance error (LDE)~\cite{wei2019regression}, max shift distance error (MSDE)~\cite{wei2019regression}, hausdorff distance (HD)~\cite{wang2020segmentation} and average symmetric surface distance (ASD)~\cite{wang2020segmentation}.

\subsubsection{Implementation details.}
For data pre-processing, each CT slice is resampled to uniform resolution ($0.5 \times 0.5 \text{mm}^{2}$), aligned by the pose rectification network and then center cropped into a patch with the size of $400 \times 304$ and $400 \times 336$ for the CQ dataset and our inhouse dataset respectively. Random horizontal flipping is applied as cheap data augmentation. The proposed model is implemented in Pytorch. We use Adam to train the model by setting  $\beta_1$ = 0.9, $\beta_2$ = 0.99 with an initial learning rate of $1e^{-3}$. The poly learning rate policy is employed. The batch size for training is set to 24, and the maximum number of epochs is set to 200. In Eq. (\ref{total}), we set $ \lambda = \gamma = \xi = 1$ and $\mu = 0.5$. And the margin $\delta$ in Eq.(\ref{connect}) is set to 1.  Moreover, the results and training details of the pose rectification network are presented in the supplementary material.
\subsubsection{Effect of Context-Aware Refinement Network.} 
\begin{figure}[t]
\includegraphics[width=\textwidth]{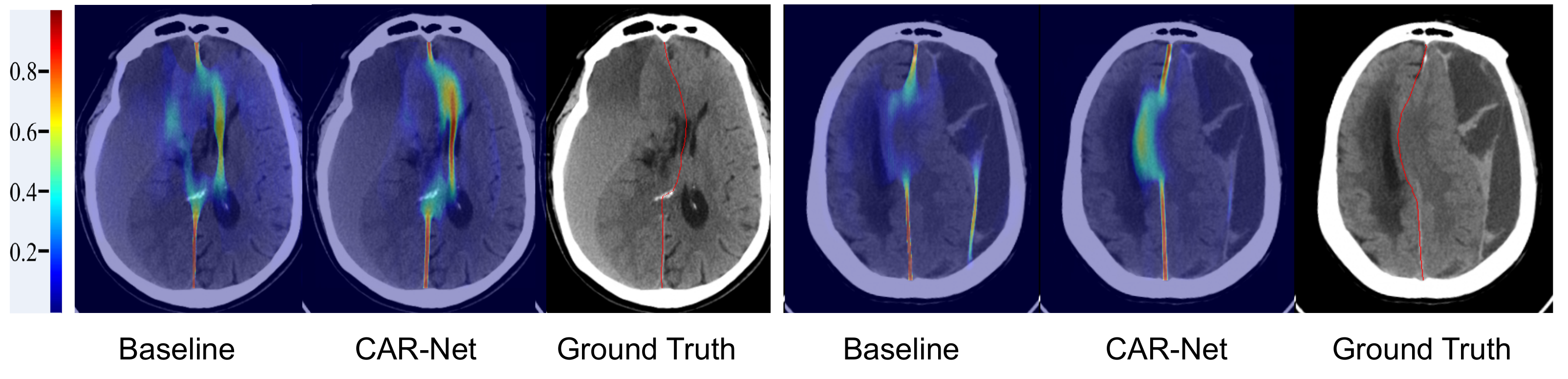}
\caption{Qualitative comparison results of segmentation probability maps between the baselinhttps://www.overleaf.com/project/5e4120603cca900001191c74e model and proposed CAR-Net.}
\label{heatmap_carnet}
\end{figure}

We replace the CAR-Net with plain U-Net in our pipeline as the baseline model. In order to obtain more contextual features, we attach a context-aware refine module based on the feature pyramid generated by the U-Net. For verifying the effectiveness of the CAR-Net, we perform ablation study on proposed CAR-Net under two loss conditions, one is training with CRL and the other is training without CRL. As shown in the last four rows of Table.~\ref{compare_result}, under both loss conditions, we observe that CAR-Net yields better performance consistently in four evaluation metrics on both datasets, compared to the baseline model. As shown in Fig.~\ref{heatmap_carnet}, the segmentation probability map of midline generated by the CAR-Net is more accurate, especially in shifted parts of largely deformed midline. The quantitative and qualitative results demonstrate that our proposed CAR-Net can obtain more contextual features, which can predict the largely deformed midline better.

\subsubsection{Effect of Connectivity Regular Loss.}
To verify the effectiveness of the proposed CRL, we conduct experiments with the baseline model and CAR-Net. It could be observed from the last four rows of Table.~\ref{compare_result} that employing CRL achieves better performance compared to the model without CRL supervision, especially in the MSDE and HD metric, which indicates the proposed CRL can reduce the error of maximum shift significantly. Furthermore, in the inference stage, the CRL of the predicted midline can also serve as a connectivity indicator to verify the performance gain of the structural connectivity. As shown in Table.~\ref{connect_result}, the connectivity indicator of the model with CRL is far smaller than counterpart without CRL. In summary, the proposed CRL can improve not only the distance performance of the midline delineation but also the midline structural connectivity effectively. Some qualitative comparisons are shown in Fig.~\ref{connect}, which further demonstrated the effectiveness of the CRL.

\begin{figure}[h]\label{connect}
\includegraphics[width=\textwidth]{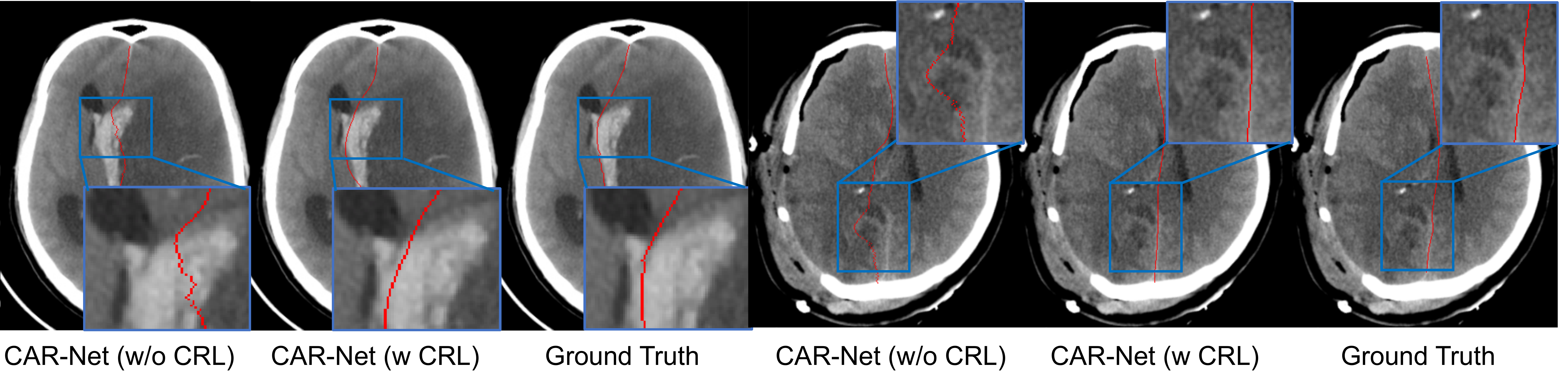}
\caption{Qualitative comparison between the CAR-Net with or without CRL.}
\end{figure}

\begin{table}[h]
\centering  
\caption{Quantitative results of the connectivity indicator in terms of mean(std) on the inhouse dataset and the CQ500 dataset.}  
\label{connect_result}
 \begin{tabular}{l|c|c|c|c}  
     \hline
        Method&\multicolumn{2}{c|}{In-house Dataset}&\multicolumn{2}{c}{CQ Dataset}\\
        \cline{2-5}
        &w/o CRL&w CRL&w/o CRL&w CRL\\
        \hline
       Baseline&0.10(1.20)&0.02(0.16)&1.12(3.72)&0.08(1.06)\\
       CAR-Net&0.41(1.86)&0.00(0.00)&0.34(1.92)&\textbf{0.01(0.12)}\\
       \hline
   \end{tabular}
\end{table}

\subsubsection{Comparisons to State-of-the-Art. }

\begin{figure}[t]
\includegraphics[width=\textwidth]{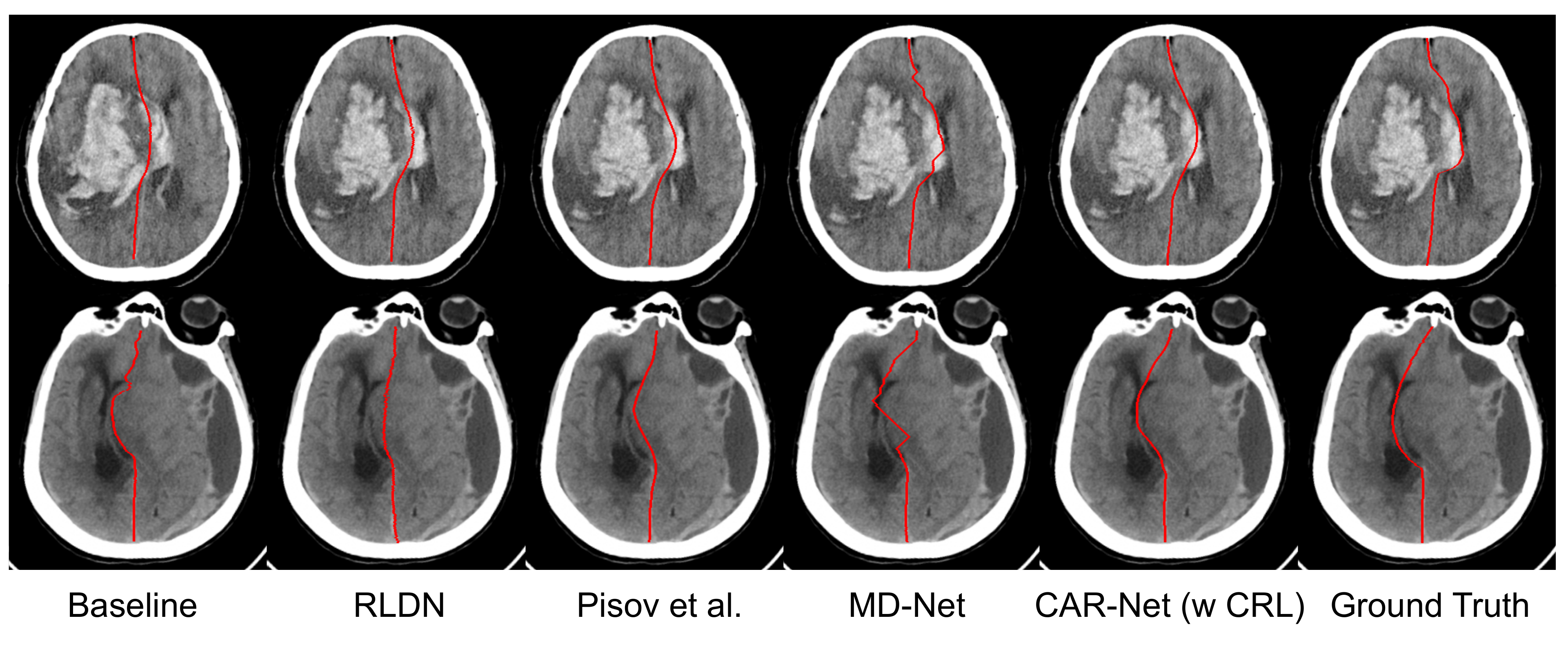}
\caption{Qualitative comparison between baseline, RLDN, Pisov et al., MD-Net and CAR-Net with CRL, showing two examples of the midline delineation.}
\label{fig_result}
\end{figure}

We provide qualitative and quantitative comparisons to three state-of-the-art algorithms of brain midline delineation: RLDN~\cite{wei2019regression}, Pisov et al.~\cite{pisov2019incorporating} and MD-Net~\cite{wang2020segmentation} on our inhouse dataset and the CQ dataset. All the experiments take the aligned image $I_{A}$ as input for fair comparison. As shown in Table.~\ref{compare_result}, our proposed model performs better than all the three methods in four evaluation metrics on both datasets, only except the comparable ASD on the inhouse dataset with the MD-Net. The experiment shows the good generalization capability and promising effectiveness of our proposed method. Fig.~\ref{fig_result} shows some delineation results of the challenging deformed brain midline. It can be inferred that our proposed method can delineate a more accurate and smoother midline, compared to the other methods, which can provide more accurate clinical judgement of pathological deformation of brain. Furthermore, the parameters of our proposed model are 3.90M, fewer than the ones of the other three methods, which can meet the needs of practical application better.

\begin{table}[h]
\centering  
\caption{Quantitative results on the inhouse dataset and CQ dataset. "*" means that MD-Net is combined with a post-processing stage, which is not an end-to-end method.}  
\label{compare_result}
 \begin{tabular}{l|c|c|c|c|c|c|c|c|c}  
     \hline
        Method& \# Params &\multicolumn{4}{c|}{Inhouse Dataset}&\multicolumn{4}{c}{CQ Dataset}\\
        \cline{3-10}
        &&LDE&MSDE&HD&ASD&LDE&MSDE&HD&ASD\\
        \hline
       RLDN~\cite{wei2019regression} &7.95 M&1.60&4.35&3.62&1.51&1.58&4.36&3.65&1.53\\
       Pisov et al.~\cite{pisov2019incorporating} &11.57 M&1.43&3.77&3.40&1.41&0.98&3.02&2.60&0.96\\
       MD-Net*~\cite{wang2020segmentation}   &4.31 M&1.10&3.49&2.94&\textbf{1.06}&1.02&3.50&2.90&0.99\\
       \hline
       Baseline(w/o CRL)&3.84 M&1.63&4.27&3.90&1.59&1.15&3.74&3.07&1.10\\
       Baseline(w CRL)&3.84 M&1.45&4.02&3.56&1.42&1.05&3.33&2.83&1.03\\
     \hline
       CAR-Net(w/o CRL)&3.90 M&\textbf{1.08}&3.24&2.84&\textbf{1.06}&0.90&3.14&2.54&0.87\\
       CAR-Net(w CRL)&3.90 M&\textbf{1.08}&\textbf{3.07}&\textbf{2.70}&1.07&\textbf{0.85}&\textbf{2.78}&\textbf{2.33}&\textbf{0.84}\\
       \hline
   \end{tabular}
\end{table}

\section{Conclusions}
We propose a context-aware refinement network (CAR-Net) to explore more discriminative contextual features and larger receptive field, which is crucial for the shifted parts of largely deformed midline. Besides, a novel connectivity regular loss (CRL) is introduced to guarantee the structural connectivity. Moreover, we address the prerequisite that the brain CT image must be in the standard pose, which is ignored by previous regression-based methods. A simple pose rectification network is presented to align the source input image to the standard pose image. The proposed method is evaluated on the CQ dataset and one inhouse dataset with the results showing that our method outperforms three state-of-the-art methods with fewer parameters.

\subsubsection{Acknowledgments}
This work was supported in part by following grants, MOST-2018AAA0102004, NSFC-61625201, Key Program of Beijing Municipal Natural Science Foundation(7191003), and following institutes, Center on Frontiers of Computing Studies, Adv. Inst. of Info. Tech and Dept. of  Computer Science, Peking University.

%
%
%

\bibliographystyle{splncs04}
\bibliography{mybibliography}

\begin{thebibliography}{10}
\providecommand{\url}[1]{\texttt{#1}}
\providecommand{\urlprefix}{URL }
\providecommand{\doi}[1]{https://doi.org/#1}

\bibitem{chen2015automatic}
Chen, M., Elazab, A., Jia, F., Wu, J., Li, G., Li, X., Hu, Q.: Automatic
  estimation of midline shift in patients with cerebral glioma based on
  enhanced voigt model and local symmetry. Australasian physical \& engineering
  sciences in medicine  \textbf{38}(4),  627--641 (2015)

\bibitem{chen2010actual}
Chen, W., Najarian, K., Ward, K.: Actual midline estimation from brain ct scan
  using multiple regions shape matching. In: 2010 20th International Conference
  on Pattern Recognition. pp. 2552--2555. IEEE (2010)

\bibitem{he2016deep}
He, K., Zhang, X., Ren, S., Sun, J.: Deep residual learning for image
  recognition. In: Proceedings of the IEEE conference on computer vision and
  pattern recognition. pp. 770--778 (2016)

\bibitem{hu2018squeeze}
Hu, J., Shen, L., Sun, G.: Squeeze-and-excitation networks. In: Proceedings of
  the IEEE conference on computer vision and pattern recognition. pp.
  7132--7141 (2018)

\bibitem{liao2018brain}
Liao, C.C., Chen, Y.F., Xiao, F.: Brain midline shift measurement and its
  automation: a review of techniques and algorithms. International journal of
  biomedical imaging  \textbf{2018} (2018)

\bibitem{liao2006tracing}
Liao, C.C., Chiang, I.J., Xiao, F., Wong, J.M.: Tracing the deformed midline on
  brain ct. Biomedical Engineering: Applications, Basis and Communications
  \textbf{18}(06),  305--311 (2006)

\bibitem{liu2014automatic}
Liu, R., Li, S., Su, B., Tan, C.L., Leong, T.Y., Pang, B.C., Lim, C.T., Lee,
  C.K.: Automatic detection and quantification of brain midline shift using
  anatomical marker model. Computerized Medical Imaging and Graphics
  \textbf{38}(1),  1--14 (2014)

\bibitem{pisov2019incorporating}
Pisov, M., Goncharov, M., Kurochkina, N., Morozov, S., Gombolevsky, V.,
  Chernina, V., Vladzymyrskyy, A., Zamyatina, K., Cheskova, A., Pronin, I.,
  et~al.: Incorporating task-specific structural knowledge into cnns for brain
  midline shift detection. In: Interpretability of Machine Intelligence in
  Medical Image Computing and Multimodal Learning for Clinical Decision
  Support, pp. 30--38. Springer (2019)

\bibitem{ronneberger2015u}
Ronneberger, O., Fischer, P., Brox, T.: U-net: Convolutional networks for
  biomedical image segmentation. In: Navab, N., Hornegger, J., Wells, W.M.,
  Frangi, A.F. (eds.) Medical Image Computing and Computer-Assisted
  Intervention – MICCAI 2015. pp. 234--241. Springer (2015).
  \doi{10.1007/978-3-319-24574-4\_28}

\bibitem{wang2020segmentation}
Wang, S., Liang, K., Pan, C., Ye, C., Li, X., Liu, F., Yu, Y., Wang, Y.:
  Segmentation-based method combined with dynamic programming for brain midline
  delineation. In: 2020 IEEE 17th International Symposium on Biomedical Imaging
  (ISBI). pp. 772--776. IEEE (2020)

\bibitem{wei2019regression}
Wei, H., Tang, X., Zhang, M., Li, Q., Xing, X., Zhou, X.S., Xue, Z., Zhu, W.,
  Chen, Z., Shi, F.: Regression-based line detection network for delineation of
  largely deformed brain midline. In: Shen, D., Liu, T., Peters, T.M., Staib,
  L.H., Essert, C., Zhou, S., Yap, P.T., Khan, A. (eds.) International
  Conference on Medical Image Computing and Computer-Assisted Intervention. pp.
  839--847. Springer (2019). \doi{10.1007/978-3-030-32248-9\_93}

\bibitem{zhang2018exfuse}
Zhang, Z., Zhang, X., Peng, C., Xue, X., Sun, J.: Exfuse: Enhancing feature
  fusion for semantic segmentation. In: Proceedings of the European Conference
  on Computer Vision (ECCV). pp. 269--284 (2018)

\end{thebibliography}
\end{document}